\documentclass[10pt,twocolumn,letterpaper,english]{article}

\usepackage{cvpr}
\usepackage{times}
\usepackage{epsfig}
\usepackage{graphicx}
\usepackage{amsmath}
\usepackage{amssymb}
\usepackage[T1]{fontenc}
\usepackage[latin9]{inputenc}
\usepackage{amsbsy}
\usepackage{amstext}
\usepackage{graphicx}
\usepackage{microtype}
\usepackage[unicode=true,
 bookmarks=false,
 breaklinks=false,pdfborder={0 0 1},backref=section,colorlinks=false]
 {hyperref}
\usepackage{array}

\cvprfinalcopy 

\def\cvprPaperID{****} 

\usepackage{booktabs} 
 
\usepackage[english]{babel}

\usepackage{xspace}

\DeclareMathOperator*{\meanoperator}{mean}

\title{R2D2: Repeatable and Reliable Detector and Descriptor}

\author{%
  Jerome Revaud \qquad Philippe Weinzaepfel \qquad César De Souza \qquad Noe Pion \\ Gabriela Csurka \qquad Yohann Cabon \qquad Martin Humenberger \\[0.1cm]
  NAVER LABS Europe \\
  \tt\small firstname.lastname@naverlabs.com
}

\makeatother

\begin{document}
\maketitle

\begin{abstract}
Interest point detection and local feature description are
fundamental steps in many computer vision applications.
Classical methods for these tasks are based on a \textit{detect-then-describe} paradigm
where separate handcrafted methods are used to first identify repeatable keypoints and then represent them with a local descriptor.
Neural networks trained with metric learning losses have recently caught up with these techniques,
focusing on learning repeatable saliency maps for keypoint detection and learning descriptors at the detected keypoint locations.
In this work, we argue that salient regions are not necessarily discriminative, and therefore can harm the performance of the description.
Furthermore, we claim that descriptors should be learned only in regions for which matching can be performed with high confidence.
We thus propose to jointly learn keypoint detection and description together with a predictor of the local descriptor discriminativeness.
This allows us to avoid ambiguous areas and leads to reliable keypoint detections and descriptions.
Our \textit{detection-and-description} approach, trained with self-supervision, can simultaneously output sparse, repeatable and reliable keypoints that outperforms state-of-the-art detectors and descriptors on the HPatches dataset. 
It also establishes a record on the recently released Aachen Day-Night localization benchmark.
\end{abstract}

\begin{figure*}
\centering
\includegraphics[width=\linewidth]{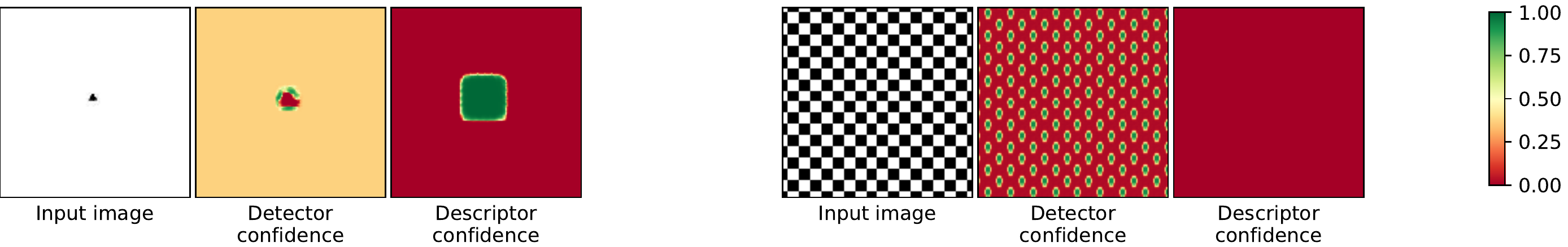}
\caption{\label{fig:toy}Toy examples to illustrate the key difference between
repeatability (2nd column) and reliability (3rd column) for a given image. Repeatable
regions in the first image are only located near the black triangle,
however, all patches containing it are equally reliable. In
contrast, all squares in the checkerboard pattern are salient hence
repeatable, but none of them is discriminative due to self-similarity.
Both confidence maps were estimated by our network.}
\end{figure*}

\section{Introduction}

Being able to accurately find and describe similar points of interest (or simply \textit{keypoints}) across images is crucial in many applications such as large-scale visual localization~\cite{svarm2016city,sattler2017large}, object detection~\cite{csurka2004visual}, pose estimation~\cite{nath2018object}, Structure-from-Motion (SfM)~\cite{schoenberger2016sfmrevisited} and 3D reconstruction~\cite{heinly2015reconstructing}.
In these applications, extracted keypoints should be sparse, repeatable, and discriminative, in order to minimize the memory footprint while maximizing matching accuracy.

Classical approaches to implement this ability are based on a two-stage pipeline that first detects keypoint locations~\cite{mikolajczyk2005comparison,harris,mikolajczyk2004scale,matas2004robust} and then computes a local descriptor for each keypoint~\cite{surf,sift}.
Specifically, the role of the keypoint detector is to look for scale-space
locations with covariance with respect to camera viewpoint changes and invariance with respect to photometric transformations. A large number of handcrafted keypoints have been
shown to work well in practice, such as corners~\cite{harris} or blobs~\cite{mikolajczyk2004scale,matas2004robust,sift}. As
for the description, various schemes based on histograms of local
gradients~\cite{brief,surf,brisk,orb}, whose most well known instance is SIFT~\cite{sift}, have been developed
and are still extensively used nowadays.

Despite this apparent success, this paradigm was recently challenged
by several approaches willing to replace the handcrafted parts by data-driven
approaches~\cite{Trzcinski2012,Mishchuk2017,Ono2018,Zieba2018,lift,matchnet,l2net,delf,savinov2017quad}.
Arguably, handcrafted methods are limited by the a priori knowledge researchers have about the tasks at hand.
The point is thus to let a deep network discover automatically which feature extraction process and representation are most suited to the data.
The few attempts for learning keypoint detectors~\cite{savinov2017quad,lfnet,lift,superpoint,d2net} have only focused on the repeatability.

On the other hand, metric learning techniques applied to learning local robust descriptors~\cite{delf,l2net} have recently outperformed traditional descriptors, including SIFT~\cite{he2018local}.
They are trained on the repeatable locations provided by the detector, which may harm the performance in regions that are repeatable but where accurate matching is not possible. Figure~\ref{fig:toy} shows such an example with a checkerboard image: every corner or blob is repeatable but matching cannot be performed due to repetitiveness of the cells.
In natural images, common textures such as the tree leafage, skyscraper windows or sea waves are also salient, but hard to match.
In this work, we claim that detection and description
are inseparably tangled since good keypoints
should not only be
\emph{repeatable} but should also be \emph{reliable} for matching.
We thus propose to learn jointly the descriptor reliability seamlessly with the detection and description processes.
Our method separately estimates a confidence map for each of these two aspects and selects only keypoints
which are both repeatable and reliable to improve the overall feature matching pipeline.

\begin{figure*}
 \centering
 \includegraphics[page=1,trim={0.1cm 18.9cm 1.6cm 0},clip,width=0.98\linewidth]{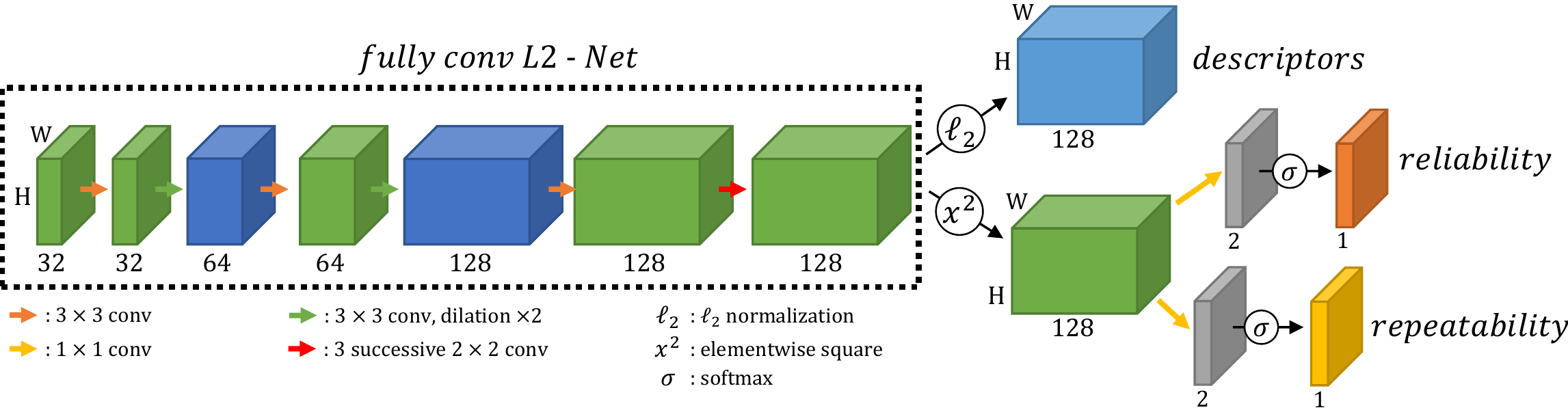}
 \caption{Overview of our network for jointly learning repeatable and reliable matches.}
 \label{fig:archi}
\end{figure*}

More precisely, our network, illustrated in Figure~\ref{fig:archi}, outputs dense local descriptors (one for each pixel) as well as two associated repeatability and reliability confidence maps.
The two maps are in fact estimates of the probabilities that a keypoint is respectively repeatable and that its descriptor is discriminative, \ie, it can be accurately matched with high confidence. 
Finally, keypoints correspond to locations that maximize both confidence maps.

To train the keypoint detector, we employ a novel unsupervised loss that encourages repeatability, sparsity as well as a uniform coverage of the image.
As for the local descriptor, it is trained with a listwise ranking loss, leveraging recent advances in metric learning based on an approximated Average Precision (AP) metric, instead of using a standard triplet or contrastive loss~\cite{Gordo2016,Schroff2015,Radenovic2016}.
We jointly learn a reliability confidence value to predict which pixels will have descriptors with a high AP, \ie, that are both discriminative, robust and in the end that can be accurately matched.
Our experiments on several benchmarks show that our formulation elegantly combines the repeatability and sparsity of the detector with a discriminative and robust descriptor.

In summary, we make three contributions:
\begin{itemize}
\item We propose a novel unsupervised loss to learn a keypoint detector: our keypoints
are sparse while still uniformly covering the image and are more repeatable than other methods.
\item A new loss to learn reliable local descriptors while explicitly estimating their reliability at the same time.
\item Our combined pipeline selects keypoints which are both repeatable and reliable and achieves state-of-the-art results.
\end{itemize}

\section{Related work}
\label{sec:related}

Local feature extraction and description play a vital role
in several high-order methods in computer vision and has received a continuous influx of attention in the
past several years (\cf surveys in  \cite{csurkaArxiv2018FromHandcrafted,gauglitz2011evaluation,salahat2017recent,tuytelaars2008local}).
Most existing works rely on a \textit{detect-then-describe} approach and we focus here on the learning approaches only.

\paragraph{Learned descriptors.}
Most deep feature matching methods have focused on learning the descriptor
component, applied either on a sparse set of keypoints~\cite{balntas2016learning,simo2015discriminative,simonyan2014learning,Mishchuk2017}
detected using standard handcrafted methods or densely over the image~\cite{fathy2018hierarchical,savinov2017matching,taira2018inloc,delf}.
The descriptor is usually trained using a metric learning loss, such
as the triplet loss~\cite{Gordo2016,Schroff2015} or a contrastive loss~\cite{Radenovic2016}.
Such loss formulation has been also used to improve descriptors for image patches~\cite{l2net,matchnet,balntas2016pn}.
Our approach has several advantages compared to these methods: 
(a) the detector is trained jointly with the descriptor, alleviating
the drawbacks of sparse handcrafted keypoint detector; 
(b) the descriptor component
is trained with an approximation of the AP loss, considering more
descriptors per batch than standard ranking losses 
(c) we jointly estimate the descriptor reliability for local feature matching.

\paragraph{Learned detectors.}
A few attempts have been recently made to learn the detector component.
The first approach for keypoint detection to rely on
machine learning was FAST~\cite{rosten2005fusing}.
Later, Di~\etal~\cite{di2018kcnn} learn to mimic the output of handcrafted detectors with a compact neural network.
In~\cite{laguna2019key}, handcrafted and learned filters are combined to detect repeatable keypoints.
These two approaches still rely on some handcrafted detectors or filters while ours is trained without such prior knowledge.
QuadNet~\cite{savinov2017quad} is an unsupervised approach based on the idea that
the ranking of the keypoint saliencies should be preserved by natural image transformations.
Following a similar approach, Zhang \etal~\cite{Zhang:2018:LTD} additionally encourage peakiness of the saliency map for keypoint detector on textures.
In this paper we employ a simpler unsupervised formulation that locally enforces the similarity of the saliency maps.

\paragraph{Jointly learned descriptor and detector.}
In the seminal LIFT approach, Yi~\etal~\cite{lift} introduced a pipeline where keypoints are detected and cropped regions
are then fed to a second network to estimate the orientation before
going throughout a third network to perform description.
Recently, the SuperPoint approach by DeTone~\etal~\cite{superpoint} 
tackles keypoint detection as a supervised task learned from
artificially generated training images containing basic structures like corners and edges.
Keypoints are then arbitrarily defined as intersection of these structures or remarkable points within, and a deep descriptor is learned jointly, sharing some of the computation. 
In contrast, our approach does not introduce a bias in the locations of 
keypoints and also does not require to compute repeatability multiple times for a given test image with different homographies.
Noh~\etal~\cite{delf} proposed DELF, an approach targeted for image retrieval that learns local features as a by-product of a classification loss coupled with an attention mechanism trained using a large-scale dataset of landmark images. 
In comparison, our approach is unsupervised and trained with relatively little data.
More similar to our approach, Mishkin \etal~\cite{mishkin2018repeatability} recently leverage deep learning to jointly enhance an affine regions detector and local descriptors. Nevertheless, their approach is rooted on a handcrafted keypoint detector that generates seeds for the affine regions, thus not truly learning keypoint detection.

More recently, D2-Net~\cite{d2net} uses a single CNN for joint detection
and description that share all weights; the detection being based
on local maxima across the channels and the spatial dimensions of the feature maps. 
Instead of arbitrarily defining keypoints as local maxima in the descriptor space, 
our approach explicitly estimates the keypoint reliability and repeatability.
Finally, Ono~\etal~\cite{lfnet} train a network from pairs of matching images 
with a complicated asymmetric gradient backpropagation scheme for the detection and a triplet loss for the local descriptor.
Compared to these works, for the first time we jointly train a sparse keypoint
detector with a deep descriptor enhanced with a reliability confidence value such that ambiguous areas are avoided.

\section{Joint learning reliable and repeatable detectors and descriptors}
\label{sec:method}

The proposed approach, referred to as R2D2, aims to predict a set of
sparse locations of an input image $I$ that are repeatable and reliable
for the purpose of local feature matching. In contrast to classical
approaches, we make an explicit distinction between repeatability
and reliability, see Figure~\ref{fig:toy}. We claim that they are in fact two complementary
aspects that must be predicted separately.

We thus propose to train a fully-convolutional network (FCN) that predicts 3 outputs for an image $I$ of size $H \times W$.
The first one is a 3D tensor $\boldsymbol{X}\in\mathbb{R}^{H\times W\times D}$ that corresponds to a set of dense D-dimensional, one per pixel.
The second one is a heatmap $\boldsymbol{S}\in[0,1]^{H\times W}$ whose goal is to provide
sparse and repeatable keypoint locations. To achieve sparsity, we
only extract keypoints at locations corresponding to local maxima
in $\boldsymbol{S}$, while $\boldsymbol{S}$ is trained to contain strong and repeatable local maxima.
The third output is an associated reliability map $\boldsymbol{R}\in[0,1]^{H\times W}$ that indicates the estimated
reliability, \ie, discriminativeness, of descriptor $\boldsymbol{X}_{ij}$ at each pixel $(i,j)$, with $i=1..W$ and $j=1..H$.

The network architecture is shown in Figure~\ref{fig:archi}. The backbone is inspired by L2-Net~\cite{l2net}. 
Compared to L2-Net, we replace the last $8\times8$ convolutional layer by 3 $2\times2$ convolutional layers, allowing to reduce the number of weights by a factor $5$ for a similar or slightly better accuracy.
The 128 dimensional output tensor serves as input to: (a) a $\ell_2$-normalization layer to obtain descriptors $\boldsymbol{X}$, (b) an element-wise square operation followed by a $1 \times 1$ convolutional layer and a softmax to obtain the reliability confidence value $\boldsymbol{R}$ of each descriptor, and (c) the same operations to obtain the repeatability map $\boldsymbol{S}$.
We now explain how we design the losses for training the network to obtain sparse, repeatable and reliable keypoints.

\subsection{Learning repeatability}

As observed in previous works~\cite{superpoint,lift}, keypoint repeatability
is a problem that cannot be tackled by standard supervised training.
In fact, using supervision essentially boils down in this case to
copying an existing detector rather than discovering better and easier
keypoints. We thus treat the repeatability as a self-supervised task
and train the network such that the positions of local maxima
in $\boldsymbol{S}$ are covariant to natural image
transformations like viewpoint or illumination changes.

Let $I$ and $I'$ be two images of the same scene and let $U\in\mathbb{R}^{H\times W\times2}$
be the ground-truth correspondences between them. In other words,
if the pixel $(i,j)$ in the first image $I$ corresponds to pixel $(i',j')$
in the second image $I'$, then $U_{ij}=(i',j')$. In practice, $U$
can be estimated using existing optical flow or stereo matching algorithms
if $I$ and $I'$ are natural images or can be obtained exactly if $I'$
was generated synthetically with a known transformation, 
\eg an homography~\cite{superpoint}, see Section~\ref{sub:train}. 
Let $\boldsymbol{S}$ and $\boldsymbol{S'}$ be the repeatability map for image $I$ and $I'$ respectively, and $\boldsymbol{S}'_{U}$ the heatmap from image $I'$ warped according to $U$.

Ultimately, we want to enforce the fact that all local maxima in $\boldsymbol{S}$
correspond to the ones in $\boldsymbol{S}'_{U}$. Our key idea is
to maximize the cosine similarity, denoted as $cosim$ in the following, between $\boldsymbol{S}$ and $\boldsymbol{S}'_{U}$.
When $cosim(\boldsymbol{S},\boldsymbol{S}'_{U})$ is maximized, 
the two heatmaps are indeed identical and their maxima correspond exactly.
However, this process assumes no occlusions, warp artifacts or border
effects which strongly impacts performance in practice. We
fix it by reformulating this idea \emph{locally}, \ie, averaging
the cosine similarity over many small patches. We define the set of
overlapping patches $\mathcal{P}=\left\{ p\right\} $ that contains
all $N\times N$ patches in $[1..W]\times[1..H]$ and define the loss as:

\begin{equation}
\mathcal{L}_{cosim}(I,I',U)=1-\frac{1}{\left|\mathcal{P}\right|}\sum_{p\in\mathcal{P}}cosim\left(\boldsymbol{S}\left[p\right],\boldsymbol{S}'_{U}\left[p\right]\right)~~,
\end{equation}
where $\boldsymbol{S}\left[p\right]\in\mathbb{R}^{N^{2}}$ denotes
the vectorized (flattened) $N\times N$ patch $p$ extracted from
$\boldsymbol{S}$, and likewise for $\boldsymbol{S}'_{U}\left[p\right]$.

Note that $\mathcal{L}_{cosim}$ can be minimized trivially by having
$\boldsymbol{S}$ and $\boldsymbol{S}'_{U}$ constant. To avoid that,
we employ a second loss function that tries to maximize the local
peakiness of the repeatability map:

\begin{equation}
\mathcal{L}_{peaky}(I)=1-\frac{1}{\left|\mathcal{P}\right|}\sum_{p\in\mathcal{P}}\left(\max_{(i,j)\in p}\boldsymbol{S}_{ij}-\meanoperator_{(i,j)\in p}\boldsymbol{S}_{ij}\right)~~.
\end{equation}

Interestingly, this allows to choose the frequency of local maxima by varying
the patch size $N$.
Finally, the resulting repeatability loss is composed
as a weighted sum of the first loss and second loss applied to both
images as:

\begin{eqnarray}
\mathcal{L}_{rep}(I,I',U) & = & \mathcal{L}_{cosim}(I,I',U) \nonumber \\
                          & + &\lambda\left(\mathcal{L}_{peaky}(I)+\mathcal{L}_{peaky}(I')\right)~~.
\label{eqn:rep}
\end{eqnarray}

\subsection{Learning reliability}

To enforce reliability, our network not only computes the repeatability map $\boldsymbol{S}$ but jointly extracts dense local descriptors
$\boldsymbol{X}$ and predicts for each descriptor $\boldsymbol{X}_{ij}\in\mathbb{R}^{D}$,
a confidence value $\boldsymbol{R}_{ij}\in[0,1]$ that estimates its reliability, \ie, discriminativeness.
The goal is to let the network learn to choose between making descriptors as discriminative as possible with a high confidence,
or a low confidence in which case the loss will have low impact on the descriptor, such as for regions that cannot be made discriminative enough.

The descriptor matching problem can be seen as
a ranking optimization problem, \ie, given two images $I$ and $I'$,
each descriptor from $I$ is searched in $I'$ as a query, ranking
all descriptors from $I'$ by increasing distance. Ranking losses
have thus been extensively and successfully used to train local descriptors
(\eg~triplet loss \cite{choy2016universal,balntas2016learning,kumar2016learning,l2net,Mishchuk2017,zhang2017learning}). At the exception of \cite{he2018local}, only
pairwise ranking losses such as the triplet loss have been used. These
losses only perform local optimization, based on a pair, triplet,
or quadruplet of training samples, which does not necessarily correlate
well with a global metric like the Average Precision (AP).
Recent work~\cite{he2018local} suggested that directly optimizing the
AP for patch descriptor matching significantly improves the performance. 
Inspired by recent advances in listwise losses \cite{He2018b,Ustinova2016}, He~\etal\cite{he2018local} defined a differentiable approximation of the AP, a standard ranking metric, that can be directly optimized during training. Given a batch of ground-truth pairs of image patches, they use a convolutional neural network to compute their descriptors.
They then compute the matrix of Euclidean distances between all patch descriptors
from the batch. Each row in the matrix can be interpreted as the distances
between a query patch from the first image and all patches from the
second image, acting as database documents. Training thus consists
in maximizing the AP computed for each query $q$ in the batch $B$ and averaged over
the whole batch.
\begin{equation}
\mathcal{L}_{AP}=\frac{1}{B}\sum_{q=1}^{B}\mathcal{L}_{AP}(q),\quad \mathcal{L}_{AP}(q)=1-AP(q)~~.
\label{eq:ap}
\end{equation}

In this work, we follow a similar path. A major difference
is that a standard keypoint detector is employed in~\cite{he2018local} to extract
patches, while our input is simply an image. The used L2-Net architecture~\cite{l2net} is
applied patch by patch, which is quite slow. Applying this network
in a fully-convolutional way is significantly more efficient. In our case, each pixel
$(i,j)$ from the first image defines a patch of size $M$ that we
can compare to all other patches in the second image. Knowing the
ground-truth correspondence $U$, we can compute its AP, which is
similar to the previous loss.

As a matter of fact, local descriptors can be extracted everywhere,
but not all locations are equally interesting. In particular, uniform
regions or elongated 1D patterns are known to lack the distinctiveness
necessary for feature matching \cite{grauman2011visual}. More interestingly,
even well textured regions are also known to be unreliable from their
semantic nature, such as tree leafages or ocean waves. It becomes
thus clear that forcefully optimizing the patch descriptor even in
meaningless regions of the image could hinder the training and runtime
performance. We therefore propose a new loss to spare the network
in wasting its efforts on undistinctive regions as:

\begin{equation}
\mathcal{L}_{AP\kappa}(i,j)=1-\big[ AP(i,j)\boldsymbol{R}_{ij}+\kappa(1-\boldsymbol{R}_{ij}) \big]~~,
\end{equation}
where $\kappa\in[0,1]$ is a hyperparameter that indicates the minimum
expected $AP$ per patch. To minimize $\mathcal{L}_{AP\kappa}(i,j)$, the
network should ideally predict $\boldsymbol{R}_{ij}=0$ if $AP(i,j)<\kappa$
and $\boldsymbol{R}_{ij}=1$ conversely. In practice, $\boldsymbol{R}_{ij}$
is between $0$ and $1$ and reflects the confidence of the network
with respect to the reliability of patch $i,j$. We found that $\kappa=0.5$
yields good results in practice. Note that a similar idea of jointly
training the descriptor and an associated confidence was recently
proposed in \cite{novotny2018self}. However, they used a triplet loss, not an AP
loss, which prevents the use of an interpretable threshold $\kappa$ as
in our case.

\noindent \textbf{Runtime.}
At test time, we run the trained network multiple times on the input image
at different scales starting from the original scale, and downsampling by $2^{1/4}$ each time until the image is smaller than 128 pixels. For each scale, we find local
maxima in $\boldsymbol{S}$ and gather descriptors from $\boldsymbol{X}$ at corresponding locations. Finally, we keep a shortlist of the best $K$ descriptors over all scales where
the descriptor score is computed as product $\boldsymbol{S}_{ij}\boldsymbol{R}_{ij}$, \ie, requiring high values for both repeatability and reliability.

\subsection{Training details}
\label{sub:train}

\noindent \textbf{Training data.}
For training, the loss needs to be computed at potentially any image location as we do not know the salient regions in advance.
To generate dense ground-truth matches, we consider two solutions:
(a) using a pair of images where the second one is obtained by applying a known transformation to the first image (homographic transform, color jittering, etc.)~\cite{lfnet}; (b) using a pair coming from an image sequence or a set of unordered images.
In the latter case, in contrast to some previous work that focused on points verified by Structure-from-Motion (SfM), we designed a pipeline based on optical flow tools that can reliably extract dense correspondences
given one image pair and a few sparse SfM-verified correspondences.
As a first step, we run a SfM pipeline~\cite{schoenberger2016sfmrevisited} that outputs a list of 3D points
and 6D camera pose corresponding to each image.
For each image pair with a sufficient overlap (\ie, with some common 3D points),
we then compute the fundamental matrix. We found that computing
the fundamental matrix directly from the 2D SfM correspondences is
more reliable than directly using the 6D camera pose.
Next, we compute high-quality dense correspondences using EpicFlow~\cite{epicflow}. We enhance the method by adding epipolar constraints in DeepMatching~\cite{deepmatching}, the first step of EpicFlow that produces semi-sparse matches.
In addition, we also predict a mask where the flow is reliable. Optical flow is
by definition defined everywhere, even in occluded areas. However,
we can obviously not train from these areas. We post-process the output
of Deep Matching as follows: we compute a graph of connected consistent neighbors,
and keep only matches belonging to large connected components (at
least 50 matches). The mask is defined using a thresholded kernel density estimator on the verified matches.
In practice, we use pairs of randomly transformed images from the distractors added recently to the Oxford and Paris retrieval dataset~\cite{radenovic2018revisiting}, that are basically images from the web. We also use pairs extracted (with the help of SfM) from the Aachen Day-Night dataset~\cite{Sattler2018CVPR,Sattler2012BMVC} which contains images from the old inner city of Aachen, Germany.

\noindent \textbf{Training sampling for AP loss.}
To have a setup as realistic as possible, given hardware constraints,
we sub-sample query pixels (in the first image) on a regular grid of $8 \times 8$ pixels from cropped images of resolution $192 \times 192$.
In the second image, we consider corresponding pixels of the queries as well as pixels sampled on a regular grid with a step of $8$ pixels.
To handle the inherent imperfection of flow and matches, we define the positives as the pixels within a radius of 4 pixels from the optical flow precision, and the negatives as all pixels at more than 8 pixels distance form this position.

\noindent \textbf{Training parameters.}
We optimize the network using Adam with a batch size of $8$, a learning rate of $0.001$ and weight decay of $0.0005$.

\section{Experimental results}
\label{sec:exps}

\subsection{Dataset and metrics}\label{sec:datasets}

We evaluate our method in the 116 full image sequences of the HPatches dataset \cite{balntas2017hpatches}.
The HPatches dataset contains 116 scenes where the first image is taken as a reference and subsequent images in a sequence are used to form pairs with increasing difficulty.
This dataset can also be further separated into 57 sequences containing large changes in illumination and 59 with large changes in viewpoint.

\noindent \textbf{Repeatability.}
Following~\cite{mikolajczyk2004scale}, we compute the repeatability
score for a pair of images as the number of point correspondences
found between the two images divided by the minimum number of keypoint
detections in the image pair. We report the average
score over all image pairs.

\noindent \textbf{Matching score (M-score).}
We follow the definitions given in \cite{superpoint,lift}.
The matching score is the average ratio between ground-truth correspondences
that can be recovered by the whole pipeline and the total number of estimated features within the shared
viewpoint region when matching points from the first image
to the second and the second image to the first one.

\noindent \textbf{Mean Matching Accuracy (MMA).}
We use the same definition as in~\cite{d2net} where the MMA score is the average percentage of correct
matches in an image pair considering multiple pixel error thresholds.
We report the average score for each threshold over all image pairs.

\subsection{Impact of repeatability patch size}

\begin{figure*}
\resizebox{\linewidth}{!}{
\begin{tabular}{>{\Huge}c>{\Huge}c>{\Huge}c}
\includegraphics[width=1\linewidth]{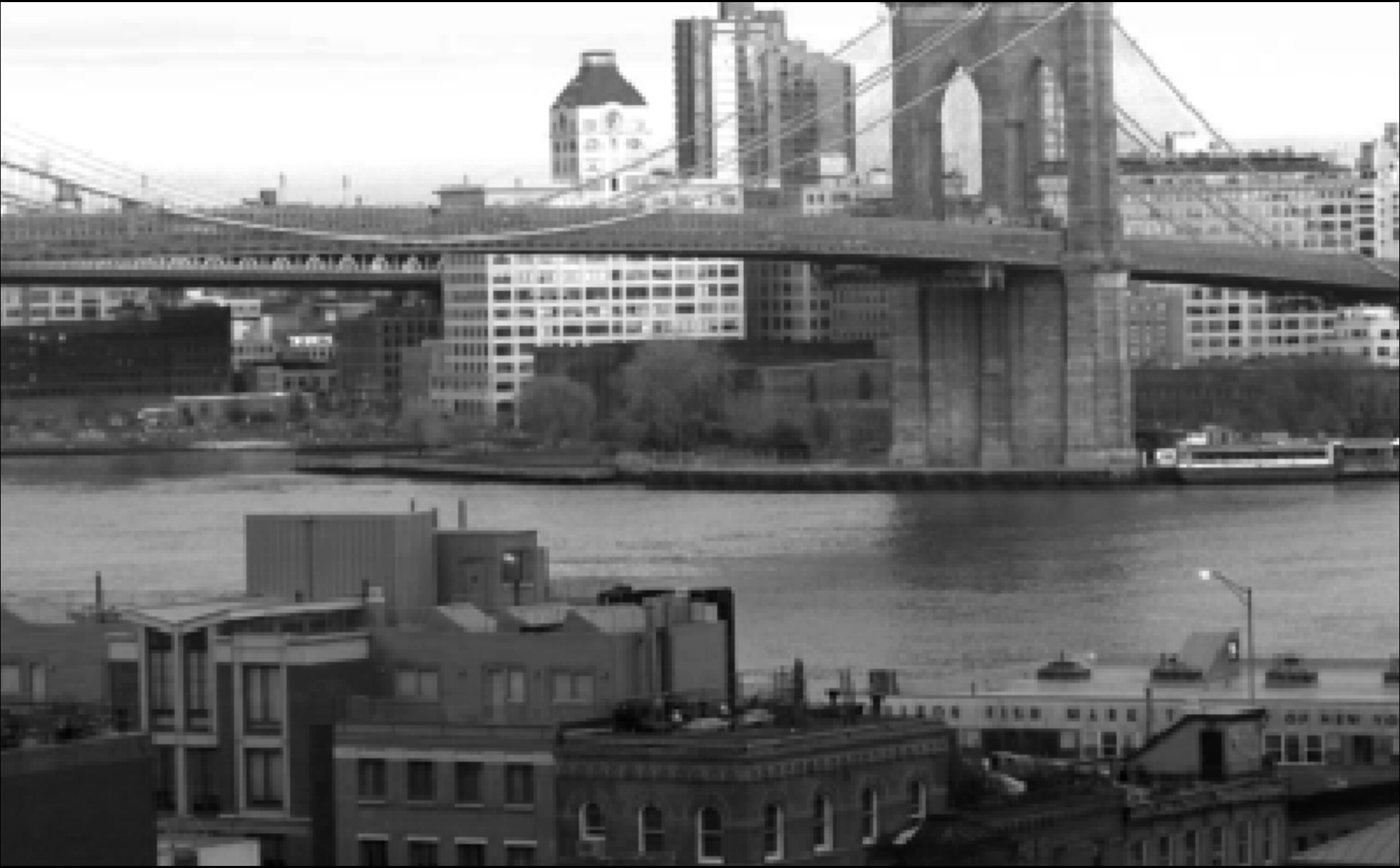} & \includegraphics[width=1\linewidth]{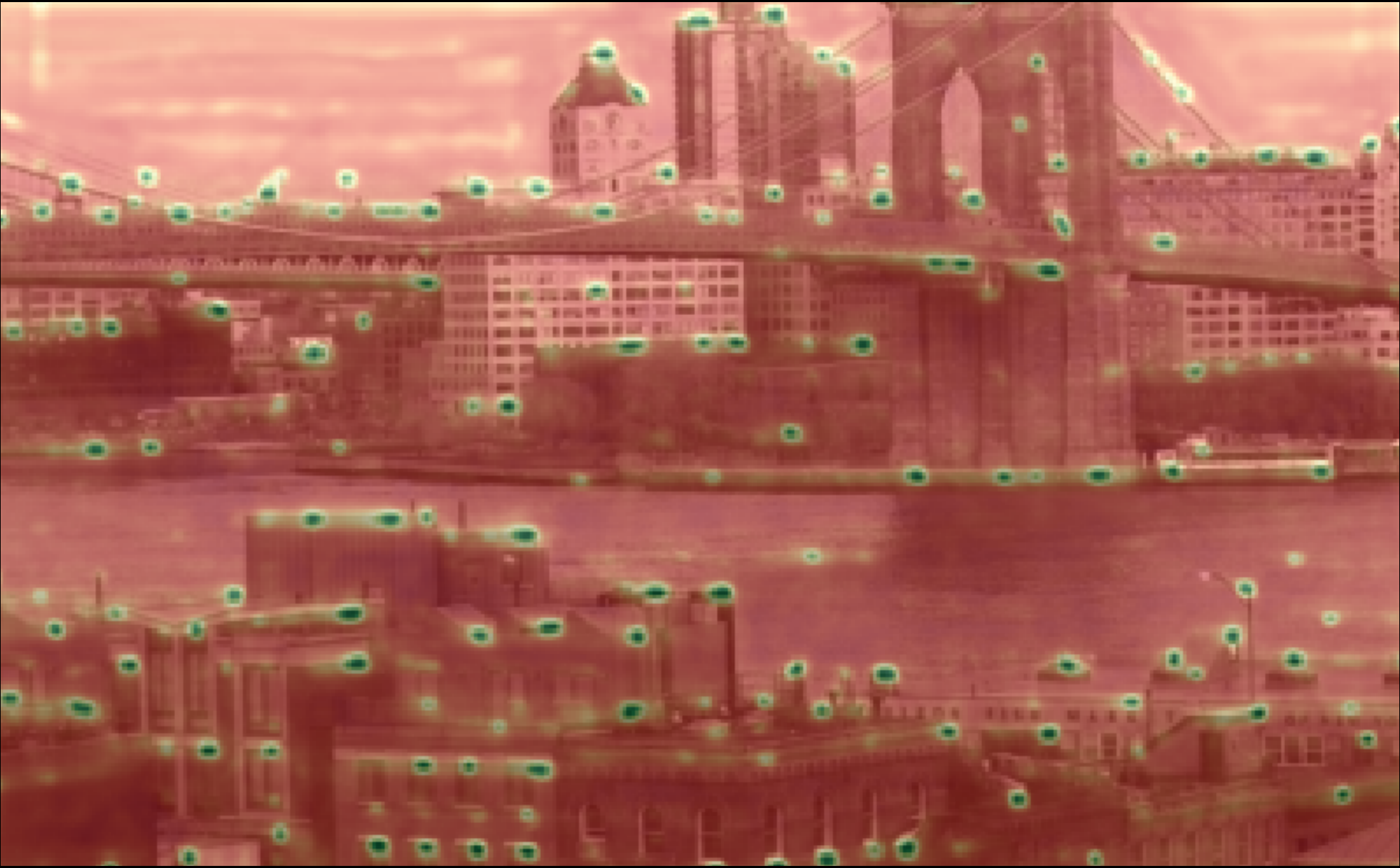} & \includegraphics[width=1\linewidth]{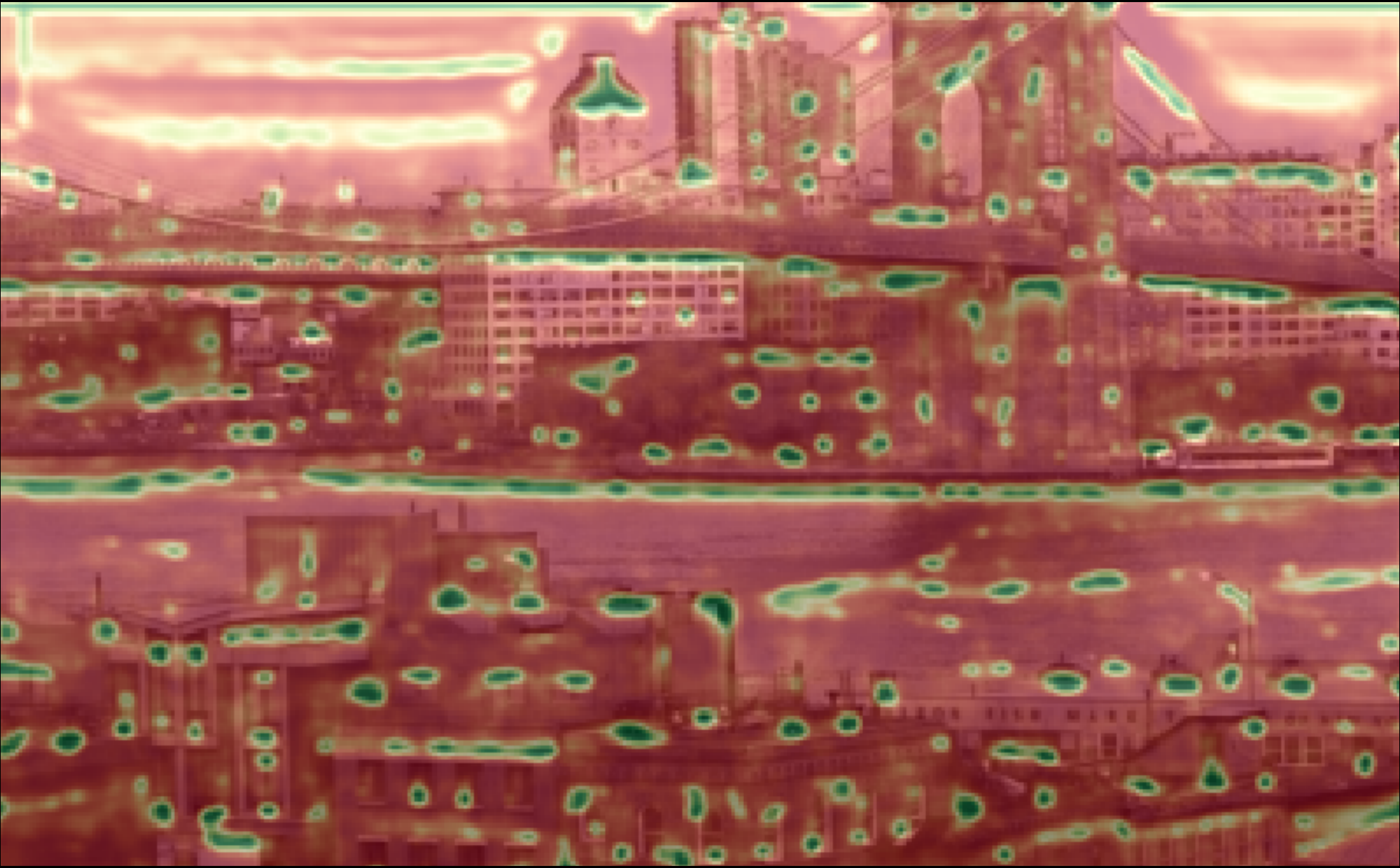}\tabularnewline
(a) input image & (b) Repeatability heatmap $\boldsymbol{S}$ for $N=64$ & (c) Repeatability heatmap $\boldsymbol{S}$ for $N=32$\tabularnewline
\includegraphics[width=1\linewidth]{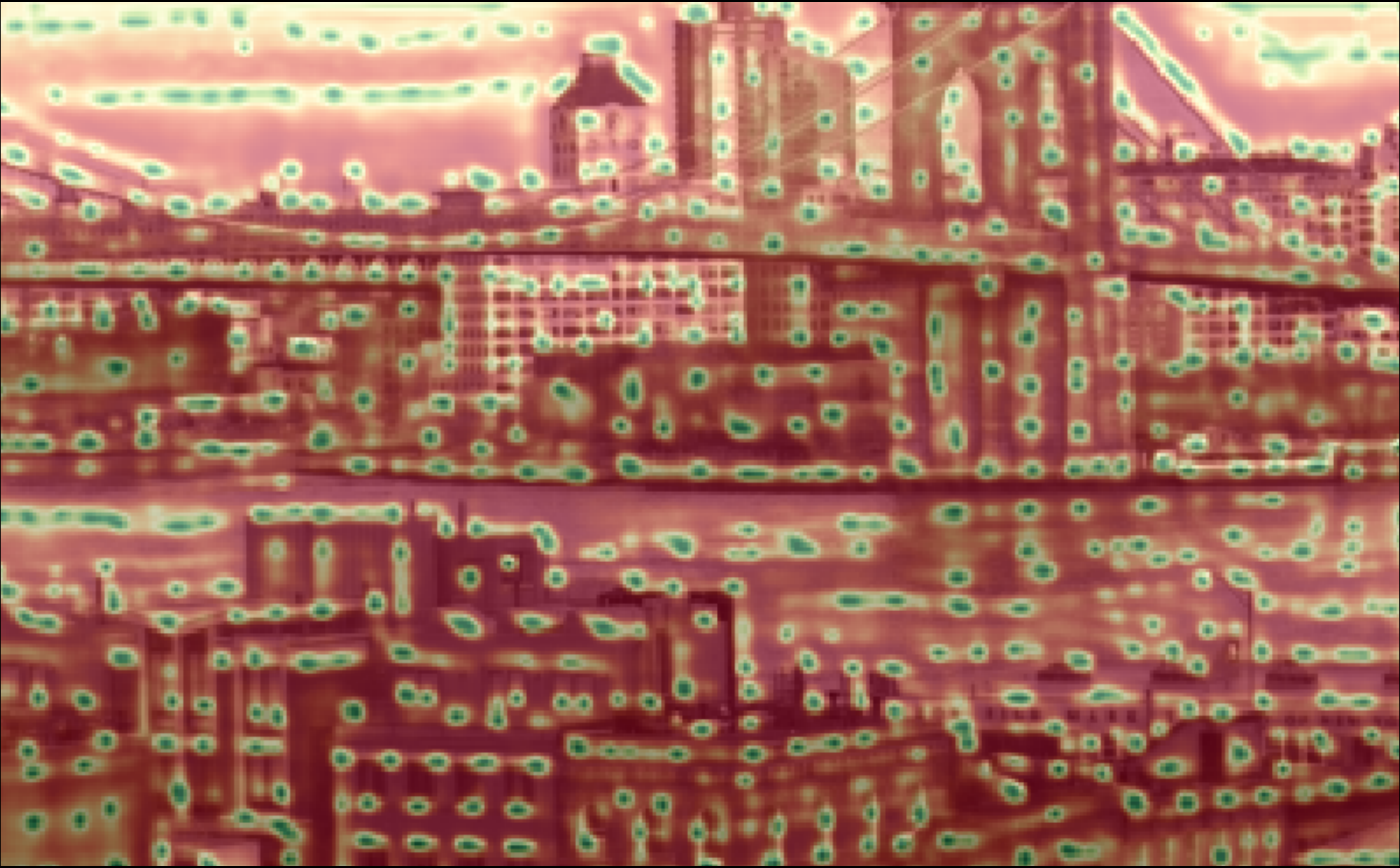} & \includegraphics[width=1\linewidth]{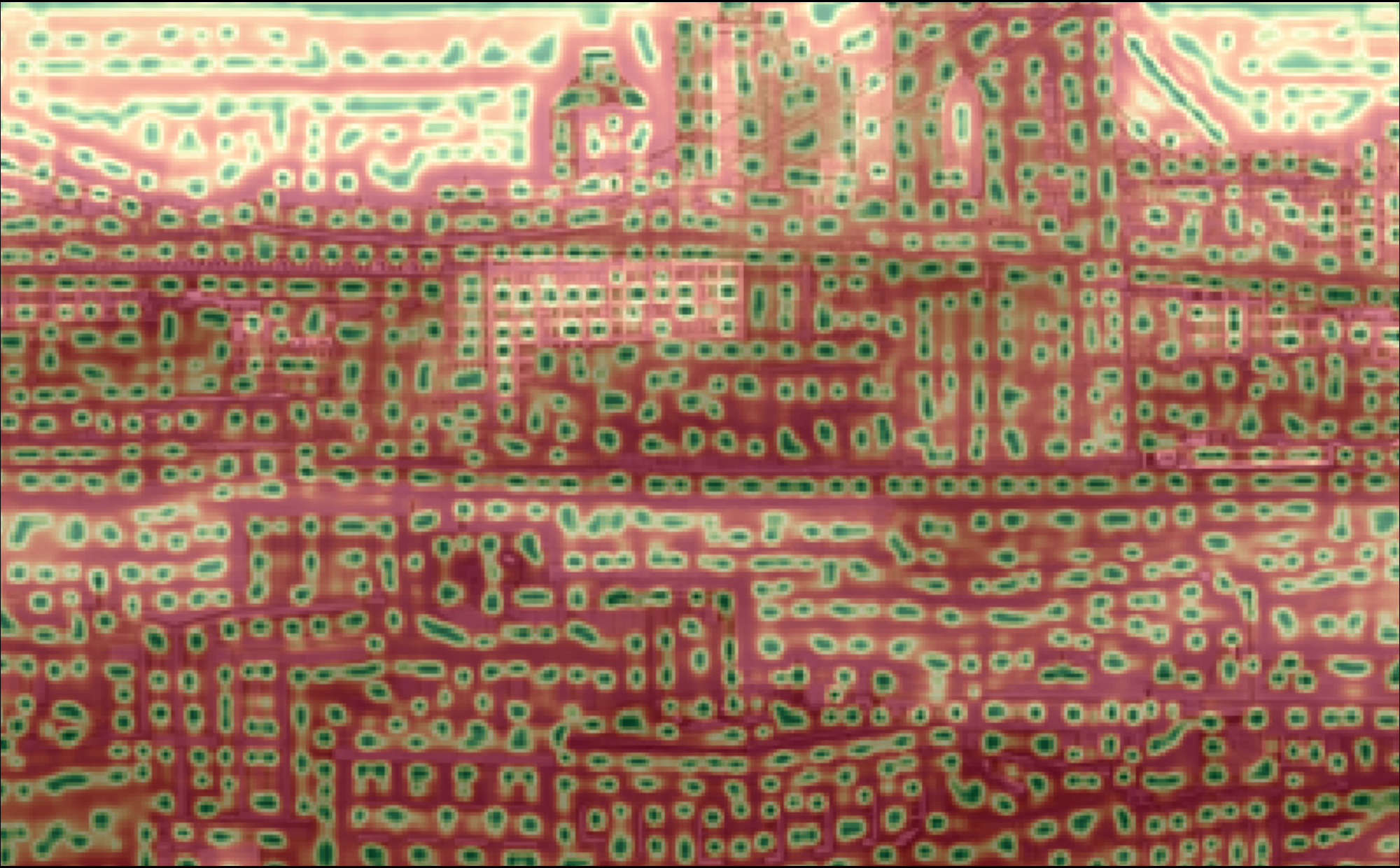} & \includegraphics[width=1\linewidth]{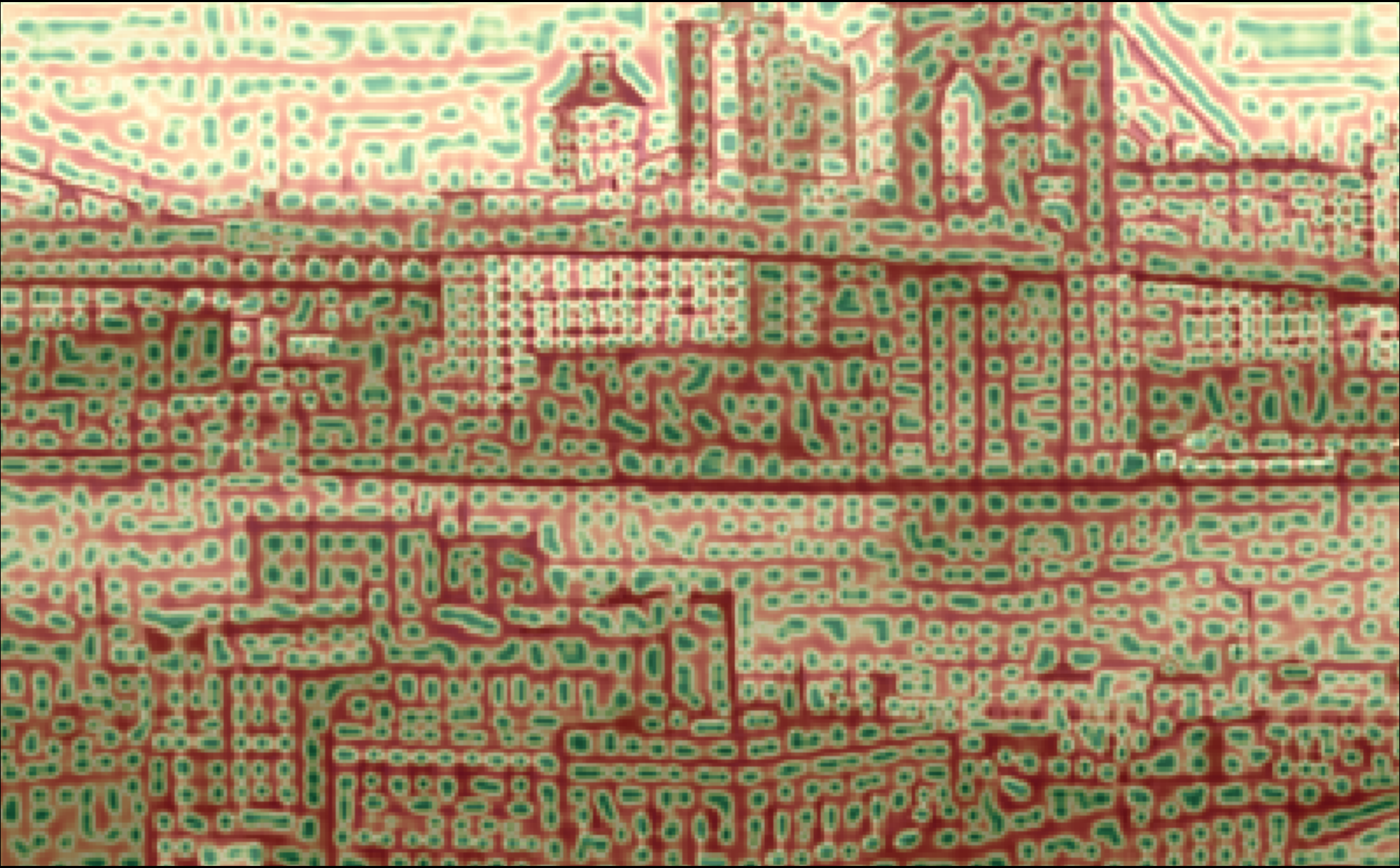}\tabularnewline
(d) Repeatability heatmap $\boldsymbol{S}$ for $N=16$ & (e) Repeatability heatmap $\boldsymbol{S}$ for $N=8$ & (f) Repeatability heatmap $\boldsymbol{S}$ for $N=4$\tabularnewline
\end{tabular}
}
\caption{
Sample repeatability heatmaps obtained when training the repeatability
losses $\mathcal{L}_{peaky}$ and $\mathcal{L}_{rep}$ with different
patch size $N$. Red and green colors denote low and high values, respectively.}
\label{fig:repmap}
\end{figure*}

\begin{figure}
 \centering
 \includegraphics[width=0.9\linewidth]{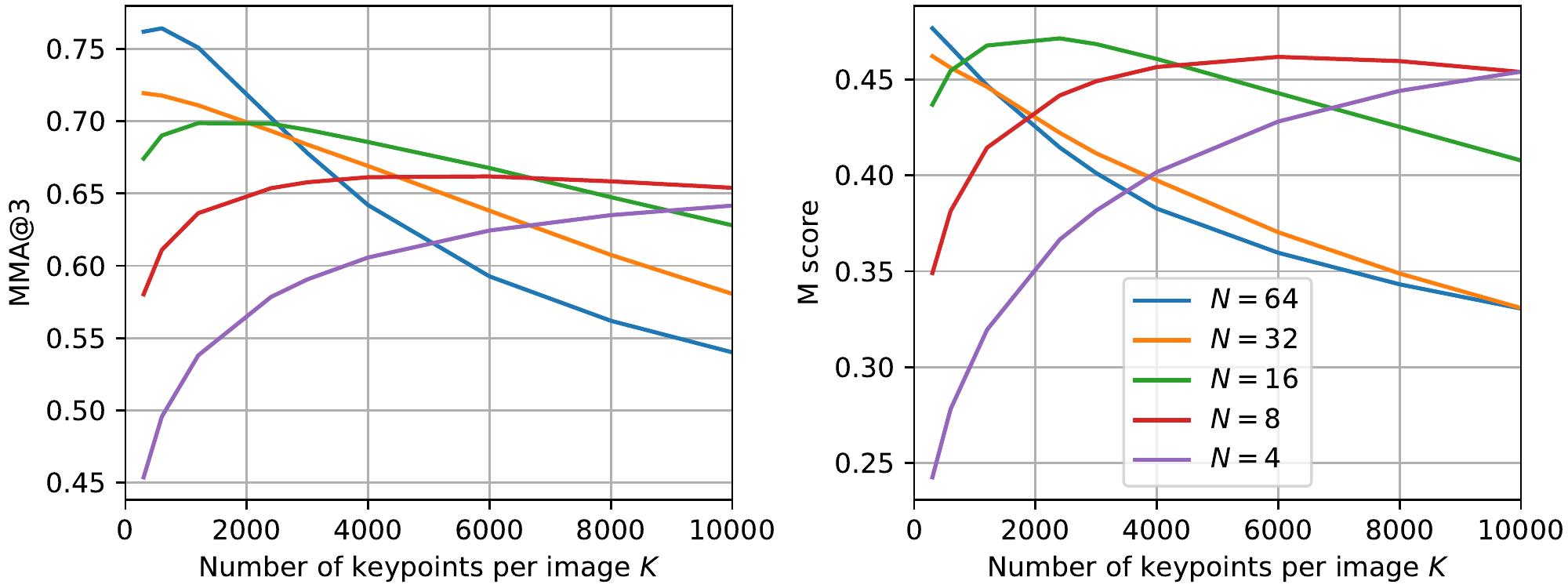}
 \caption{
  MMA@3 and M-score for different patch sizes $N$ on the HPatches dataset,
  as a function of the number of retained keypoints $K$ per image.
 }
 \label{fig:kpts}
\end{figure}

We first evaluate the impact of the patch size $N$ used in the repeatability loss $\mathcal{L}_{rep}$, see Equation~\ref{eqn:rep}.
The local patch size essentially controls the number of keypoints as the loss ideally encourages the network to output a single local maxima per window of size $N \times N$.
We show in Figure~\ref{fig:repmap} different repeatability maps obtained 
from the same input image for networks trained with different values of $N$.
We observe that when $N$ is large, the network outputs few highly-repeatable keypoints, and conversely for smaller values of $N$.
Note that the networks even learn to populate empty regions like the sky
with a grid-like pattern when $N$ is small, while it avoids them when $N$ is large.

We also plot the mean matching accuracy on the HPatches dataset in Figure~\ref{fig:kpts}
for various $N$ as a function of the number of retained keypoints $K$ per image.
As expected, models trained with large $N$ strongly outperforms models with lower $N$
when the number of retained keypoints is low, since as seen above these keypoints have a higher quality. 
When keeping more keypoints, poor local maxima starts to get selected for these models
(\eg in the sky in Figure~\ref{fig:repmap}(b)) and the matching performance drops.
However, having numerous keypoints is important for many applications such as visual localization because it augments the chance that at least a few of them will be correctly matched despite occlusions or other noise sources.
There is therefore a trade-off between the number of keypoints and the matching performance.
In the following experiments, and unless stated otherwise, we use a model trained with $N=16$ and $K=5000$ keypoints per image.

\begin{figure}
 \centering
 \includegraphics[width=0.9\linewidth]{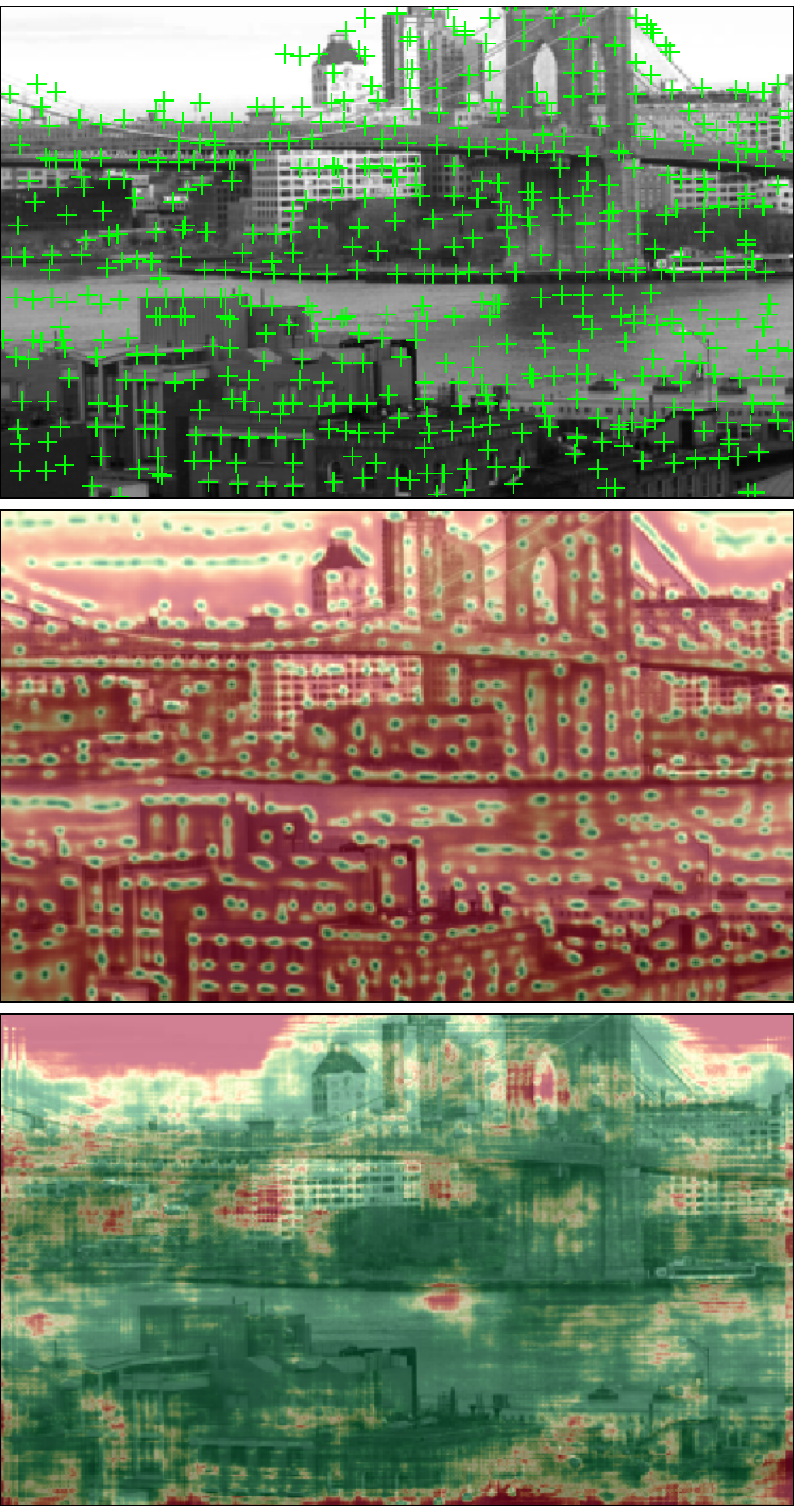}
 \caption{For one given input image (1st row), we show the repeatability (2nd row)
  and reliability heatmaps (3rd row) extracted at a single scale,
  overlaid onto the original image.
  Valid keypoints (both repeatable and reliable) are shown as crosses
  in the first image.
 }
 \label{fig:heatmaps}
\end{figure}

\begin{table}
\centering
\begin{tabular}{cc||cc}
\toprule 
reliability & repeatability & M-score & MMA@3\\
\midrule
$\checkmark$ &  & 0.304 & 0.512\\
 & $\checkmark$ & 0.436 & 0.680\\
$\checkmark$ & $\checkmark$ & \textbf{0.461} & \textbf{0.686}\\
\bottomrule
\end{tabular}
\caption{\label{tab:abla}Ablative study on HPatches. We report the M-score and the MMA at a 3px error threshold for our method (bottom row) as well as our approach without repeatability map (top row) or reliability map (middle row).}
\end{table}

\subsection{Impact of separate reliability and repeatability}

Our main contribution is to show that repeatability and reliability can be predicted separately and help to jointly learn detector and descriptor.
In Table~\ref{tab:abla}, we report the performance when removing the repeatability $\boldsymbol{S}$, \ie, keypoints are defined by maxima of the reliability map, or removing the reliability map $\boldsymbol{R}$, \ie, learning the descriptor with the AP loss formulation of Equation~\ref{eq:ap} on all pixels.
Without repeatability, the performance significantly drops both in terms of MMA@3 and M-score. This shows that repeatability is not well correlated with the descriptor reliability.
When training without estimating the descriptor reliability, the M-score decreases by 3\% and the MMA@3 by 0.6\%. This shows the importance of estimating the discriminativeness of descriptors.
In Figure~\ref{fig:heatmaps} we show the repeatability and reliability heatmaps obtained for the input image. Despite its small size, the network was able to learn that the sky region is a region that cannot be matches. More interestingly, more complex patterns are also rejected, such as 1-d patterns (under the bridge) or grid patterns (building windows). As a result, keypoints in those regions are scored low and are not retained in the top-$K$ final output (see top row of Figure~\ref{fig:heatmaps}).

\subsection{Comparison with the state of the art}

We now compare our approach to state-of-the-art keypoint detectors and descriptors on the HPatches and Aachen datasets.

\noindent \textbf{Detector repeatability.}
We first evaluate the keypoints extracted by our approach in term of repeatability.
Following~\cite{savinov2017quad}, we report the repeatability on the Oxford dataset~\cite{mikolajczyk2005comparison}, a subset of HPatches, for which the transformations applied to sequences is known and include jpeg compression (JPEG), blur (Blur), zoom and rotation (Z+R), luminosity (L), and viewpoint perspective (VP).
Table~\ref{tab:repeat} shows a comparison with QuadNet~\cite{savinov2017quad} and the handcrafted Difference of Gaussians (DoG) used in SIFT~\cite{sift} on this dataset when varying the number of interest points.
We observe that overall our approach significantly outperforms these two approaches, in particular for a high number of interest points. 
This demonstrates the excellent repeatability of our detector.
Note that training on the Aachen dataset may obviously helps for street views. 
Nevertheless, our approach performs well even for the cases of blur or rotation (bark, boat), while we did not train the network for such challenging cases.

\begin{table*}
\centering
\small
\resizebox{0.72\linewidth}{!}{
\begin{tabular}{cccccccc}
\toprule 
 &  &  & \multicolumn{5}{c}{{\small{}Number of interest points}}\tabularnewline
\midrule 
{\small{}Transformations } & {\small{}Data } & {\small{}Method } & {\small{}300 } & {\small{}600 } & {\small{}1200 } & {\small{}2400 } & {\small{}3000}\tabularnewline
\midrule 
{\small{}Viewpoint Perspective (VP) } & {\small{}graf } & {\small{}DoG } & 0.21 & 0.0.2 & 0.18 & - & -\tabularnewline
 &  & {\small{}QuadNet } & 0.17 & 0.19 & 0.21 & 0.24 & 0.25\tabularnewline
 &  & {\small{}Ours } & \textbf{0.32} & \textbf{0.38} & \textbf{0.42} & \textbf{0.45} & \textbf{0.47}\tabularnewline
\midrule 
 & {\small{}wall } & {\small{}DoG } & 0.27 & 0.28 & 0.28 & - & -\tabularnewline
 &  & {\small{}QuadNet } & 0.3 & 0.35 & 0.39 & 0.44 & 0.46\tabularnewline
 &  & {\small{}Ours } & \textbf{0.62} & \textbf{0.62} & \textbf{0.65} & \textbf{0.70} & \textbf{0.71}\tabularnewline
\midrule 
{\small{}Zoom and Rotation (Z+R)} & {\small{}bark } & {\small{}DoG } & 0.13 & 0.13 & - & - & -\tabularnewline
 &  & {\small{}QuadNet } & 0.12 & 0.13 & 0.14 & 0.16 & 0.16\tabularnewline
 &  & {\small{}Ours } & \textbf{0.27} & \textbf{0.33} & \textbf{0.37} & \textbf{0.44} & \textbf{0.47}\tabularnewline
\midrule 
 & {\small{}boat } & {\small{}DoG } & 0.26 & 0.25 & 0.2 & - & -\tabularnewline
 &  & {\small{}QuadNet } & 0.21 & 0.24 & 0.28 & 0.28 & 0.29\tabularnewline
 &  & {\small{}Ours } & \textbf{0.33} & \textbf{0.39} & \textbf{0.45} & \textbf{0.54} & \textbf{0.57}\tabularnewline
\midrule 
{\small{}Luminosity (L) } & {\small{}leuven } & {\small{}DoG } & 0.51 & 0.51 & 0.5 & - & -\tabularnewline
 &  & {\small{}QuadNet } & \textbf{0.7} & \textbf{0.72} & \textbf{0.75} & \textbf{0.76} & \textbf{0.77}\tabularnewline
 &  & {\small{}Ours } & 0.65 & 0.69 & 0.73 & \textbf{0.76} & \textbf{0.77}\tabularnewline
\midrule 
{\small{}Blur (B) } & {\small{}bikes } & {\small{}DoG } & 0.41 & 0.41 & 0.39 & - & -\tabularnewline
 &  & {\small{}QuadNet } & 0.53 & 0.53 & 0.49 & 0.55 & 0.57\tabularnewline
 &  & {\small{}Ours } & \textbf{0.66} & \textbf{0.67} & \textbf{0.71} & \textbf{0.75} & \textbf{0.76}\tabularnewline
\midrule 
 & {\small{}trees } & {\small{}DoG } & 0.29 & 0.3 & 0.31 & - & -\tabularnewline
 &  & {\small{}QuadNet } & \textbf{0.36} & \textbf{0.39} & 0.44 & 0.49 & 0.5\tabularnewline
 &  & {\small{}Ours } & 0.28 & 0.36 & \textbf{0.45} & \textbf{0.55} & \textbf{0.6}\tabularnewline
\midrule 
{\small{}Compression (JPEG) } & {\small{}ubc } & {\small{}DoG } & \textbf{0.68} & 0.6 & - & - & -\tabularnewline
 &  & {\small{}QuadNet } & 0.55 & \textbf{0.62} & \textbf{0.66} & \textbf{0.67} & \textbf{0.68}\tabularnewline
 &  & {\small{}Ours } & 0.40 & 0.45 & 0.54 & 0.65 & \textbf{0.68}\tabularnewline
\bottomrule
\end{tabular}
}
\caption{\label{tab:repeat}Comparison with QuadNet~\cite{savinov2017quad} and a handcrafted difference of gaussian (DoG) in terms of detector repeatability on the Oxford dataset.}
\end{table*}

\begin{figure*}
\centering
\includegraphics[width=\linewidth]{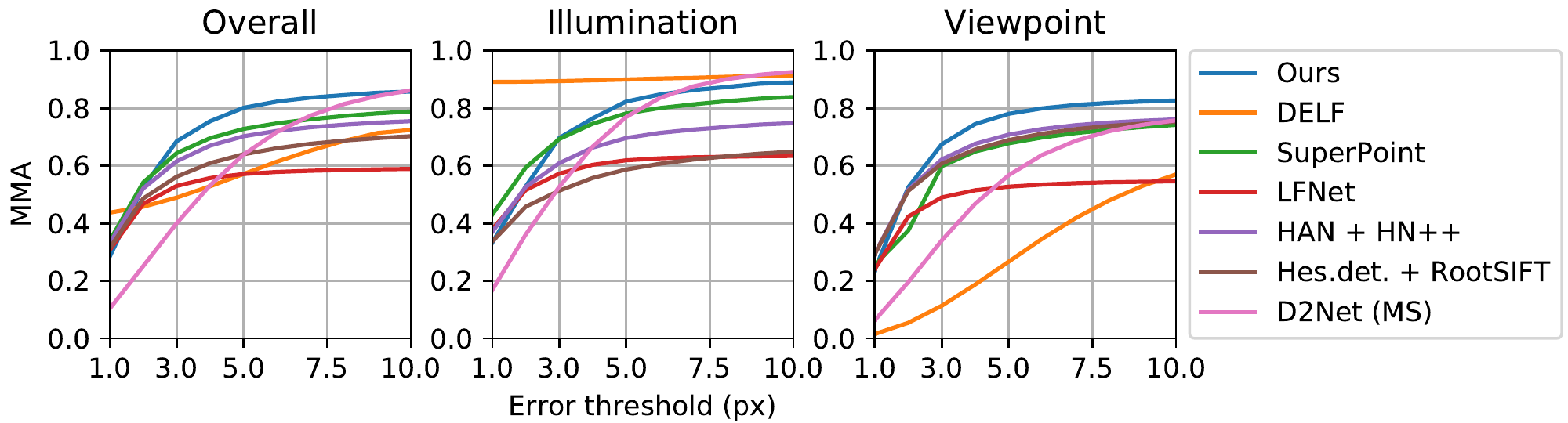}
\caption{Comparison with the state of the art using the MMA for varying error thresholds on the HPatches dataset.}
\label{fig:sota}
\end{figure*}

\noindent \textbf{Mean Matching Accuracy.}
We next compare the mean matching accuracy with the state of the art, namely DELF~\cite{delf}, SuperPoint~\cite{superpoint}, LF-Net~\cite{lfnet}, multi-scale D2-Net~\cite{d2net}, HardNet++ descriptors with HesAffNet regions~\cite{Mishchuk2017,mishkin2018repeatability} (HAN + HN++) and a handcrafted Hessian affine detector with RootSIFT descriptor~\cite{perd2009efficient}.
Figure~\ref{fig:sota} shows the results for illumination and viewpoint changes, as well as the overall performance on the HPatches dataset. 

We observe that our method significantly outperforms the state of the art in particular for middle range thresholds.
This is at the exception of DELF for illumination changes, which can be explained by the fact that they use a fixed grid of keypoints while this image subset has no spatial changes.
Interestingly, our method significantly outperforms jointly detector and descriptor such as D2-Net~\cite{d2net}, in particular at low level thresholds, which mean that our keypoints benefit from our joint training with repeatability and reliability.

\noindent \textbf{Matching score.}
At 3px error threshold, we obtain a M-Score of $0.425$ compared to $0.335$ reported by LF-Net~\cite{lfnet} and $0.288$ for SIFT~\cite{sift}. This demonstrates again the benefit of our matching approach with repeatability and reliability.

\begin{figure*}
\includegraphics[width=0.49\linewidth]{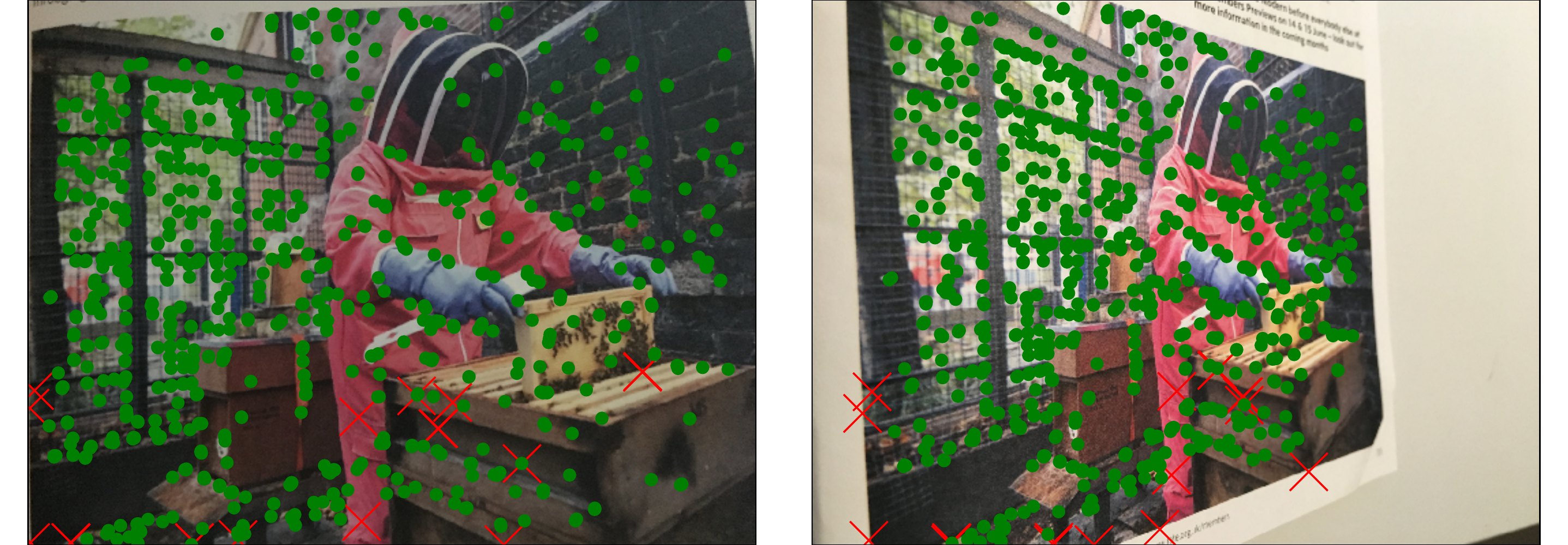}
\includegraphics[width=0.49\linewidth]{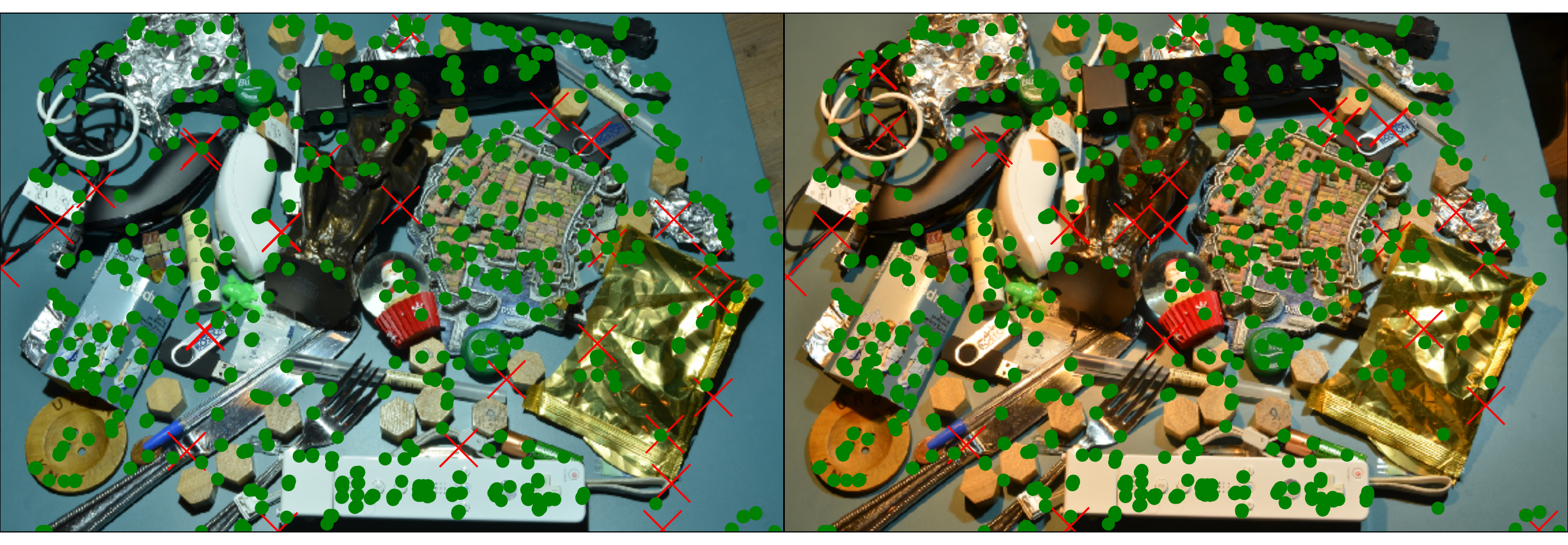}
\caption{\label{fig:ex-MMA}Sample results using reciprocal nearest matching.
Correct and incorrect correspondences are shown as green dots and red
crosses, respectively.}
\end{figure*}

\noindent \textbf{Qualitative results.}
We show in Figure~\ref{fig:ex-MMA} two examples of matching pair with a drastic change of viewpoint (left) and illumination change (right). We observe that our matches cover the entire image and most of them are correct (green dots).

\subsection{Applications to visual localization}

In this section, we evaluate our method for the task of visual localization~\cite{svarm2016city,sattler2017large}, where the goal is to estimate the camera position within a given environment using an image. This is particularly interesting because robust local feature matching is crucial to enable visual localization in real-world conditions where it faces challenges such as day-night transitions and significant viewpoint changes between training and testing.
First, we present a comparison with state of the art methods. Second, we present an ablative study in order to show the impact of training data.

\paragraph{Localization pipeline.}
The evaluation is done using \emph{The Visual Localization Benchmark\footnote{\url{https://www.visuallocalization.net}}}, more specifically we use the local feature challenge of CVPR19. In order to evaluate a feature extraction method, a pre-defined visual localization pipeline\footnote{\scriptsize \url{https://github.com/tsattler/visuallocalizationbenchmark/tree/master/local_feature_evaluation}} 
based on COLMAP~\cite{schoenberger2016sfmrevisited} is used. First, the custom features (the ones to evaluate) are used to generate a structure-from-motion model. Second, the test images are registered in this model again using the custom features. For feature matching, mutual nearest neighbor is used. Everything else follows COLMAP. The pipeline generates three result numbers representing the percentages of successfully localized images within three error tolerances $(0.5m, 2\deg)$ / $(1m, 5\deg)$ / $(5m, 10\deg)$, where the first number represents the max.~position error in meters and the second number represents the max.~orientation error in degrees. The dataset used is Aachen Day-Night~\cite{Sattler2018CVPR,Sattler2012BMVC}.

\begin{table*}
\centering
\begin{tabular}{cccc||cc|cc|cc}
Method & \#kpts & dim & \#weights & \multicolumn{2}{c|}{0.5m, 2$^{\circ}$} & \multicolumn{2}{c|}{1m, 5$^{\circ}$} & \multicolumn{2}{c}{5m, 10$^{\circ}$}\tabularnewline
\hline 
\hline 
RootSIFT\cite{sift} & 11K & 128 & - & 33.7 & (-12) & 52.0 & (-14) & 65.3 & (-23)\tabularnewline
HAN+HN\cite{mishkin2018repeatability} & 11K & 128 & 2M & 37.8 & (-8) & 54.1 & (-12) & 75.5 & (-13)\tabularnewline
SuperPoint\cite{superpoint} & 7K & 256 & 1.3M & 42.8 & (-3) & 57.1 & (-9) & 75.5 & (-13)\tabularnewline
DELF (new)\cite{delf} & 11K & 1024 & 9M & 39.8 & (-6) & 61.2 & (-5) & 85.7 & (-3)\tabularnewline
D2-Net\cite{d2net} & 19K & 512 & 15M & 44.9 & (-1) & 66.3 & (-0) & \textbf{88.8} & (-0)\tabularnewline
\hline 
\hline 
R2D2, $N=16$ & 2.5K & 128 & 0.5M & 43.9 & (-2) & 61.2 & (-5) & 84.7 & (-4)\tabularnewline
R2D2, $N=16$ & 5K & 128  & 0.5M  & \textbf{45.9}  & (-0)  & 65.3  & (-1)  & 86.7  & (-2)\tabularnewline
R2D2, $N=16$ & 10K  & 128  & 0.5M  & 44.9 & (-1) & \textbf{67.3} & (+1) & 87.8 & (-1)\tabularnewline
R2D2, $N=8$ & 10K  & 128  & 0.5M  & 45.9 & (-0) & 63.3 & (-3) & 87.8 & (-1) \tabularnewline
R2D2, $N=8$ & 10K & 128 & 1.0M & \textbf{45.9} & - & 66.3 & - & \textbf{88.8} & -\tabularnewline
\end{tabular}
\caption{Comparison to the state of the art on the Aachen Day-Night dataset. We report the percentages of successfully localized images within 3 error thresholds (0.5m and 2$^{\circ}$, 1m and 5$^{\circ}$, 5m and 10$^{\circ}$). The number in parenthesis indicates the performance difference compared to our best model in the last row.}
\label{tab:sotaa}
\end{table*}

\paragraph{Impact of $\boldsymbol{N}$ and $\boldsymbol{K}$.} 
We have evaluated our approach in several configurations and report their performance in Table~\ref{tab:sotaa}. Namely, we have evaluated our model trained with $N=16$ for different numbers of keypoints $K$ per image.
For visual localization, it can be interesting to output more  keypoints per image as it increases the chances of having at least few keypoints correctly matched despite occlusions or strong viewpoint/illumination changes, which in turn improves the localization accuracy.
We therefore also evaluate our approach keeping $K=10000$ keypoints per image, this time using $N=8$ as we observed a higher MMA in this range (see Figure~\ref{fig:kpts}).
For this latter model, we have also evaluated the impact of augmenting the size of the network by doubling the number of weights in the internal convolution layers.
Our approach performs well in all configurations, including in the case with only $K=2500$ keypoints per image. Quadrupling the number of keypoints leads in slightly higher localization accuracy. Augmenting the model size results in the best overall performance.

\paragraph{Comparison with the state of the art.} 
Table~\ref{tab:sotaa} also provides a comparison to other methods submitted to the visual localization benchmark.
Our augmented model for $K=10000$ and $N=8$ outperforms all competing methods by a significant margin at all error thresholds.
The recent D2-Net approach~\cite{d2net} performs almost equally with only 1\% less images localized within 0.5m.
Interestingly, even our approach with $K=5000$ performs better than most of the other methods, whereas it uses twice less keypoints per image.
This demonstrates the high quality of our keypoint detection and scoring scheme. 
Indeed good keypoints for localization are ranked higher and even a shortlist with $K=5000$ yield good results.
In addition, we note that our local features have a relatively low dimensionality with respect to the features used in the other approaches (128 instead of 256, 512 or 1024 for others).
Our network is also very compact as it contains only 1 million parameters, which is up to 15 times less than other competing learned methods.
This shows the high efficiency or our joint detector and descriptor training based on direct AP minimization with separate repeatability and reliability.

\begin{table}
\begin{centering}
\resizebox{\linewidth}{!}{
\begin{tabular}{|c|c|c|c||cc|cc|cc|}
\hline 
W & A & S & F & \multicolumn{2}{c|}{0.5m, 2$^{\circ}$} & \multicolumn{2}{c|}{1m, 5$^{\circ}$} & \multicolumn{2}{c|}{5m, 10$^{\circ}$}\tabularnewline
\hline 
\hline 
$\checkmark$ &  &  &  & 43.9 & (-2) & 61.2 & (-4) & 77.6 & (-9)\tabularnewline
\hline 
$\checkmark$ & $\checkmark$ &  &  & 42.9 & (-3) & 60.2 & (-5) & 78.6 & (-8)\tabularnewline
\hline 
$\checkmark$ & $\checkmark$ & $\checkmark$ &  & 42.9 & (-3) & 61.2 & (-4) & 84.7 & (-2)\tabularnewline
\hline 
$\checkmark$ & $\checkmark$ &  & $\checkmark$ & 43.9 & (-2) & 63.3 & (-2) & \textbf{86.7} & (-0)\tabularnewline
\hline 
$\checkmark$ & $\checkmark$ & $\checkmark$ & $\checkmark$ & \textbf{45.9} & - & \textbf{65.3} & - & \textbf{86.7} & -\tabularnewline
\hline 
\end{tabular}
}
\end{centering}
\caption{\label{tab:aabla}Ablative study in terms of training data on the
Aachen dataset. We report the percentages of successfully localized images within 3 error thresholds (0.5m and 2$^{\circ}$, 1m and 5$^{\circ}$, 5m and 10$^{\circ}$). The number in parenthesis is the difference compared to the model trained on all data.
All results are presented for $N=16$ and $K=5000$ keypoints
per image. W=web images + homographies; A=Aachen-day images + homographies;
S=Aachen-day-night from automatic style transfer + homographies; F=Aachen-day
images pairs with optical flow.}
\end{table}

\paragraph{Ablative study.}
Our network is trained from image pairs gathered from 4 different sources, see Section~\ref{sub:train}, in equal proportions: random web images (W), Aachen-day images (A),
Aachen-night images obtained from automatic style transfer~\cite{LiECCV18Closedform}\footnote{We used the code provided at \url{https://github.com/NVIDIA/FastPhotoStyle} specifically the version without semantic segmentation.} (S). Image pairs are obtained synthetically from random homographies for W, A and S. Finally, real Aachen-day image pairs automatically annotated by computing the optical flow guided by the structure-from-motion model of the training images.
We present in Table~\ref{tab:aabla} the percentages of successfully localized images when training our networks on different subsets of the training data.
Interestingly, our method performs well even for a network trained from only web images with homographies, significantly outperforming SIFT~\cite{sift}, SuperPoint~\cite{superpoint} and the more recent HesAffNet~\cite{mishkin2018repeatability}.
Adding images from Aachen-day surprisingly does not result in any major change. This shows that our choice of a relatively small architecture prevents the network from overfitting.
Synthetically generated night images enables a significant improvement for large error thresholds. In comparison, adding real image pairs annotated with optical flow enables a larger performance boost at all error thresholds. Finally, combining all 4 training sources yields to the best performance.

\section{Conclusion}
\label{sec:conclusion}

We proposed a new learning-based feature extraction method which jointly detects and describes keypoints in images.
In contrast to traditional handcrafted features, our method learns both keypoint repeatability and a confidence for keypoint reliability from relevant training data.
Our network is trained with self-supervision using a mixture of synthetic (images with known transformations) and real data (point correspondences). 
Furthermore, we use style transfer methods to increase robustness against drastic illumination changes such as day-night transitions. 
Our experiments on the standard benchmark HPatches as well as for the task of visual localization show superior results of our approach in comparison to state-of-the-art methods. 

{\small{}\bibliographystyle{ieee}
\bibliography{biblio}
}{\small\par}
\end{document}